\pdfoutput=1
\PassOptionsToPackage{round}{natbib}
\PassOptionsToPackage{table,dvipsnames}{xcolor}
\PassOptionsToPackage{hyphens}{url}
\documentclass[11pt, logo, onecolumn, copyright, colorlinks=true, allcolors=blue]{nvidiatechreport}

\usepackage[utf8]{inputenc}
\usepackage{xcolor}
\usepackage{tcolorbox}
\tcbuselibrary{breakable,skins,listings}
\usepackage{graphicx}
\usepackage{booktabs}
\usepackage{amsmath}
\usepackage{amsfonts}
\usepackage{amssymb}
\usepackage{microtype}
\usepackage{xspace}
\usepackage{tabularx}
\usepackage{subcaption}
\usepackage{multirow}
\usepackage{makecell}
\usepackage{float}
\usepackage{stfloats}
\usepackage{placeins}
\usepackage[round]{natbib}
\usepackage{subcaption}
\usepackage{wrapfig}
\usepackage{placeins}

\definecolor{gold}{RGB}{212,175,55}
\definecolor{silver}{RGB}{192,192,192}
\definecolor{bronze}{RGB}{205,127,50}

\definecolor{pastelblue}{RGB}{173,216,230}
\definecolor{pastelyellow}{RGB}{255,253,208}
\definecolor{pastelpink}{RGB}{255,209,220}
\definecolor{pastelgreen}{RGB}{176,226,172}
\definecolor{pastellavender}{RGB}{230,230,250}
\definecolor{NvidiaGreen}{RGB}{118,185,0}

\newcommand{\answerTODO}[1][]{\textcolor{red}{\bfseries [TODO]}}
\newcommand{\justificationTODO}[1][]{\textcolor{red}{\bfseries [TODO]}}

\title{\centering{Post-Training Language Models for Gold-Medal Performance in Coding Competitions}}
\author{
Aleksander Ficek\protect\footnotemark[1],
Sean Narenthiran\protect\footnotemark[1],
Mehrzad Samadi,
Somshubra Majumdar,
Boris Ginsburg
}

\renewcommand{\thefootnote}{\fnsymbol{footnote}}

\begin{abstract}
\large \textbf{Abstract.}
Competitive programming has become a key test of large language model reasoning, with international competitions such as IOI and ICPC representing its most challenging settings. We present an end-to-end specialization pipeline combining large-scale problem curation, synthetic reasoning traces, supervised fine-tuning (SFT), and reinforcement learning (RL). Using 22,000 curated problems, we train Nemotron-3-Nano-CC (30B-A3B) with SFT and RL and Nemotron-3-Ultra-CC (550B-A55B) with SFT alone. We further introduce GenCorrect, a feedback-driven test-time compute strategy that iteratively generates, evaluates, and refines diverse solutions. On IOI 2025, Nano-CC improves from 130 points to 291 after post-training and to 468 with GenCorrect, exceeding the gold threshold of 438.3 while Ultra-CC reaches 502. Guided by these results, we develop a competition-specific Ultra-CC system and evaluate it prospectively during IOI 2026. Under the same time, internet-access, and submission constraints as human contestants, it scores 535.4 out of 600, exceeding both the gold threshold of 361.12 and the top human score of 498.27. To our knowledge, this is the first AI system to outscore the highest-scoring human contestant on an IOI problem set.

\end{abstract}

\begin{document}

\maketitle
\footnotetext[1]{Equal contribution.}
\renewcommand{\thefootnote}{\fnsymbol{footnote}}
\setcounter{footnote}{1}

\section{Introduction}

Competitive programming has emerged as a challenging domain for evaluating the reasoning capabilities of large language models (LLMs). Unlike conventional coding benchmarks, competitive programming requires models to synthesize novel algorithms, reason over complex constraints, and produce implementations that pass hidden tests under time and memory limits. Performance on competitive programming has become a strong indicator of a model's reasoning and coding ability~\citep{li2022alphacode,jain2025livecodebench,zheng2025livecodebenchpro}. Recent years have seen progress on these tasks, with proprietary and open-weight models achieving gold-medal performance in competitions such as the International Olympiad in Informatics (IOI) and the International Collegiate Programming Contest (ICPC)~\citep{openai2025competitive,deepseekai2025v32,samadi2025gencluster,yang2026nemotroncascade2}. Despite this progress, the contributions of components required to reach gold-medal performance remain difficult to isolate. Existing systems are often closed, rely on specialized models, or combine changes in training data, post-training, model scale, and inference-time compute.

\begin{figure*}[t]
    \centering
    \begin{subfigure}[t]{0.66\textwidth}
        \centering
        \includegraphics[width=\linewidth,trim={0.2cm 0.1cm 0.4cm -0.2cm}]
            {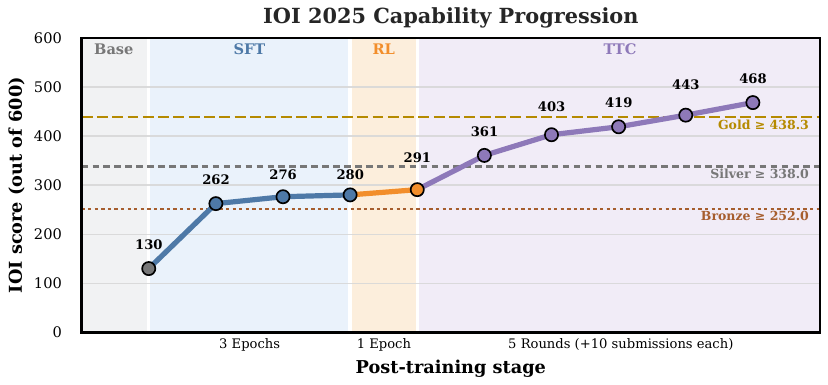}
        \caption{General Nano-CC on IOI 2025: SFT and RL improve Score@1, while GenCorrect achieves gold-medal performance with 50 submissions.}
        \label{fig:main_capability_progression}
    \end{subfigure}
    \hfill
    \begin{subfigure}[t]{0.28\textwidth}
        \centering
        \includegraphics[width=\linewidth,trim={0.4cm -0.13cm 0.3cm 0.8cm}]
            {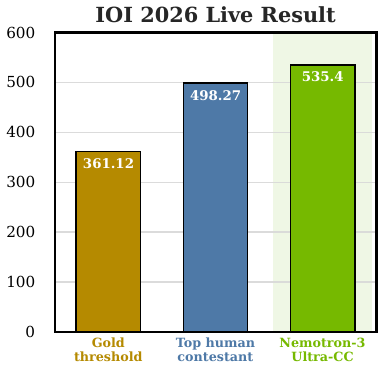}
        \caption{Competition Ultra-CC surpasses highest human score.}
        \label{fig:ioi_2026_result}
    \end{subfigure}
    \vspace{-0.2cm}
    \caption{Performance of our pipeline and models on IOI 2025 and IOI 2026.}
    \vspace{-2mm}
    \label{fig:combined_results}
\end{figure*}

We present a competitive-programming pipeline combining large-scale problem curation, synthetic data generation, supervised fine-tuning (SFT), reinforcement learning (RL), and iterative test-time refinement. We curate 22,000 problems and use DeepSeek-V4-Flash~\citep{deepseekai2026deepseekv4} to generate 1.2 million reasoning traces for our compact model and 477,642 traces for our larger model, excluding and deduplicating all evaluation problems from training. Starting from Nemotron-3-Nano-30B-A3B~\citep{nvidia2025nemotron3nanoopen}, we apply SFT and RL to produce \textbf{Nemotron-3-Nano-CC}, a 30B-parameter mixture-of-experts model with 3B active parameters. Starting from Nemotron-3-Ultra-550B-A55B~\citep{nvidia2026nemotron3ultraopen}, we apply SFT without code-specific RL to produce \textbf{Nemotron-3-Ultra-CC}, a 550B-parameter model with 55B active parameters. Both models use GenCorrect, our iterative inference strategy that refines diverse solutions using feedback from previous submissions. Figure~\ref{fig:main_capability_progression} summarizes the progression of Nemotron-3-Nano-CC on IOI 2025. Its Score@1 increases from 130 to 280 after SFT and 291 after RL, while five GenCorrect rounds raise its score to 468, exceeding the gold threshold of 438.3 under the official 50-submission limit~\citep{ioi2025results}. Our general Nemotron-3-Ultra-CC reaches 304 without code-specific RL and 502.0 after five rounds of GenCorrect. 

After completing these general-pipeline experiments, we apply their insights to develop a competition-specific model and inference pipeline for IOI 2026. We use IOI 2025 as a development benchmark to select the training and inference adaptations used by this system. We then evaluate the resulting system live on the IOI 2026 problem set under competition time and submission, before the problems are publicly released.\footnote{Our system was not an official IOI contestant and the run was not supervised by IOI. Therefore, its score was not included in the official rankings and the evaluation is reported as an unofficial, unsupervised benchmark.} Under matched competition time, internet-access, and submission constraints, Competition Ultra-CC scores 535.4/600, exceeding the gold threshold of 361.12 and the top official contestant score of 498.27, as shown in Figure~\ref{fig:ioi_2026_result}~\citep{ioi2026results}. We plan to release our competition Nemotron-3-Ultra-CC checkpoint together with runnable inference and evaluation recipes in NeMo-Skills.\footnote{\url{https://github.com/NVIDIA-NeMo/Skills}} To summarize, our main contributions are:

\begin{itemize}
    \item We develop an end-to-end specialization pipeline for competitive programming, including large-scale problem curation, synthetic reasoning data, long-context supervised fine-tuning, and reinforcement learning with executable rewards.
    \item We introduce GenCorrect, a closed-loop test-time compute strategy that iteratively generates diverse solutions, incorporates evaluator feedback, and refines subsequent generations under a constrained submission budget.
    \item We provide a comprehensive empirical analysis of how synthetic data, SFT, RL, model scale, and test-time compute contribute to competitive-programming performance.
    \item We apply our pipeline to develop a competition-specific system evaluated live during IOI 2026. Under the same time limits, submission platform, and internet restrictions as human contestants, it scored 535.4/600, surpassing the highest-scoring human contestant. To our knowledge, this is the first time an AI system has done so on any IOI problem set.
\end{itemize}

\section{Competition Settings}
\subsection{International Olympiad in Informatics.} 

IOI is an individual programming competition conducted over two contest days~\citep{ioi2026rules}. At IOI 2025, contestants were presented with six problems, each worth up to 100 points and divided into subtasks covering different input constraints. A submission received credit for each subtask whose test cases it passed, allowing partially correct or less efficient algorithms to earn partial scores. Contestants could submit at most 50 solutions per problem, and the six problem scores were summed to produce a maximum score of 600~\citep{ioi2025results,openai2025competitive}. Medal thresholds were determined from the final score distribution, with approximately the top 1/12 of contestants receiving gold, the next 1/6 receiving silver, and the next 1/4 receiving bronze~\citep{ioi2026rules}.

\subsection{International Collegiate Programming Contest.} 

At the ICPC 2025 World Finals, teams of three contestants competed for five hours using a single shared computer. The contest contained 12 algorithmic problems evaluated using binary scoring: a problem was solved only when a submission passed all hidden test cases. Teams were ranked primarily by the number of problems solved, with ties broken by total penalty time. For each solved problem, penalty time consisted of the elapsed contest time before acceptance plus 20 minutes for each preceding incorrect submission~\citep{icpcworldfinalsrules}. The top four teams received gold medals, teams placing 5th--8th received silver medals, and teams placing 9th--12th received bronze medals~\citep{icpc2025results}.

\begin{figure*}[t]
    \centering
    \includegraphics[
        width=\linewidth,
        trim={0.65cm 0.1cm 0.65cm 0.1cm}
    ]{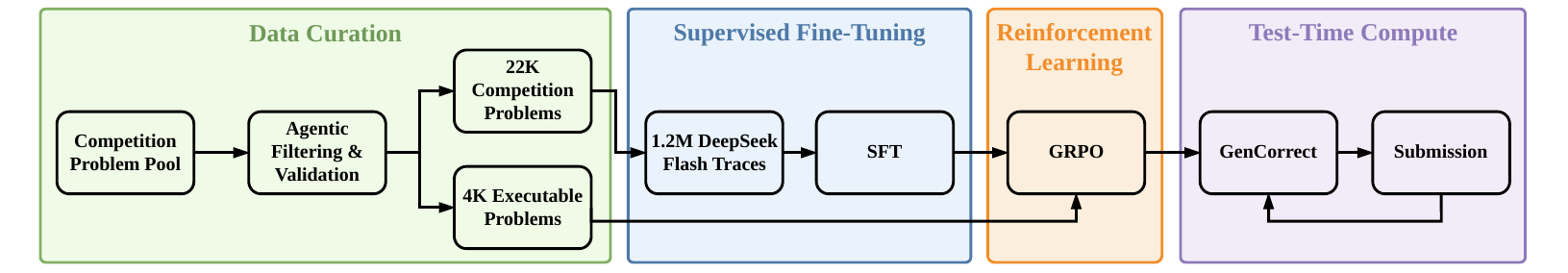}
    \caption{Our competitive-programming pipeline. Nemotron-3-Nano-CC undergoes supervised fine-tuning and reinforcement learning, while Nemotron-3-Ultra-CC uses supervised fine-tuning only. Both models use GenCorrect at inference time.}
    \label{fig:post_training}
    \vspace{-2mm}
\end{figure*}

\section{Method}

Our pipeline, summarized in Figure~\ref{fig:post_training}, combines supervised fine-tuning (SFT), reinforcement learning (RL), and iterative test-time compute. Modifications used for our live IOI 2026 benchmark run are described in Section~\ref{sec_ioi_2026}.

\subsection{Data Curation}

We curate 22,000 problems from 16 regional and international competition families spanning the last two decades, together with problems from online programming platforms. An automated pipeline packages each problem into an executable evaluation environment containing its statement, constraints, test cases, auxiliary files, and reference solutions. We retain only environments that produce consistent verdicts across reference and generated solutions. We exclude all IOI 2025, ICPC 2025, and LiveCodeBench Pro problems from the SFT and RL data and deduplicate the training corpus against these evaluations. IOI 2026 is a strictly prospective evaluation because our system was run before the problems were publicly released. Appendix~\ref{app:data_curation} provides the complete construction and filtering procedure.

\subsection{Supervised Fine-Tuning}

\begin{figure}[t]
    \centering
    \includegraphics[
        width=0.8\linewidth,
        trim={0.3cm 0.5cm 0.3cm 0.5cm}
    ]{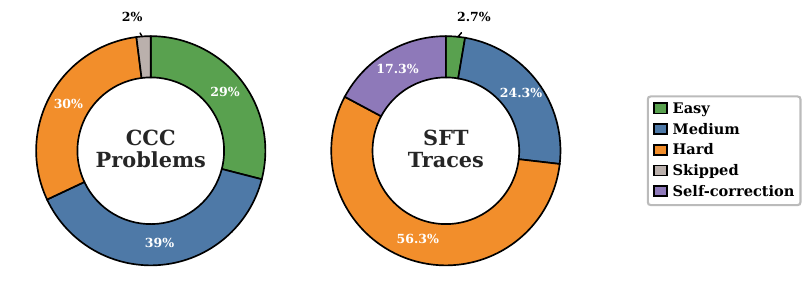}
    \caption{Composition of the curated competitive-programming corpus (left) and SFT traces (right).}
    \label{fig:data_mixtures}
\end{figure}

We use DeepSeek-V4-Flash to generate 1.2 million reasoning traces for Nemotron-3-Nano-30B-A3B and 477,642 traces for NVIDIA-Nemotron-3-Ultra-550B-A55B~\citep{deepseekai2026deepseekv4,nvidia2025nemotron3nanoopen,nvidia2026nemotron3ultraopen}. As shown in Figure~\ref{fig:data_mixtures}, we allocate more generations to difficult problems and include self-improvement traces in which the teacher refines a previously generated solution. These traces expose the models to the iterative refinement behavior used by GenCorrect. We fine-tune Nano for three epochs and Ultra for one epoch, using a global batch size of 64 and sequence packing up to 262K tokens. Ultra is initialized from its RLVR-teacher checkpoint~\citep{nvidia2026nemotron3ultraopen}. Complete optimization, parallelism, and compute settings are provided in Appendix~\ref{app:training_details}.

\subsection{Reinforcement Learning}

We apply RL only to Nemotron-3-Nano-30B-A3B. After filtering for reliable and sufficiently fast executable environments, the RL corpus contains 3,219 problems, split into 2,847 training and 372 validation problems. We split at the parent-problem level to prevent subtasks from the same problem appearing in both sets. We train using NeMo RL~\citep{nemo-rl} with Group Relative Policy Optimization (GRPO)~\citep{shao2024deepseekmathpushinglimitsmathematical}. Each step samples 16 rollouts for each of 64 prompts at temperature 1.0, yielding 1,024 rollouts. Generated C++17 solutions are compiled and executed, receiving a terminal reward of 1 for full credit and 0 otherwise. We optimize a token-level clipped policy-gradient objective with no reference-policy KL penalty~\citep{yu2025dapo} and select the final checkpoint using held-out validation performance. Further details are provided in Appendix~\ref{app:training_details}.

\subsection{Test-Time Compute}
\label{subsection:ttc}

\begin{figure*}[t]
    \centering
    \includegraphics[
        width=\linewidth,
        trim={0.6cm 0.2cm 0.6cm 0.2cm}
    ]{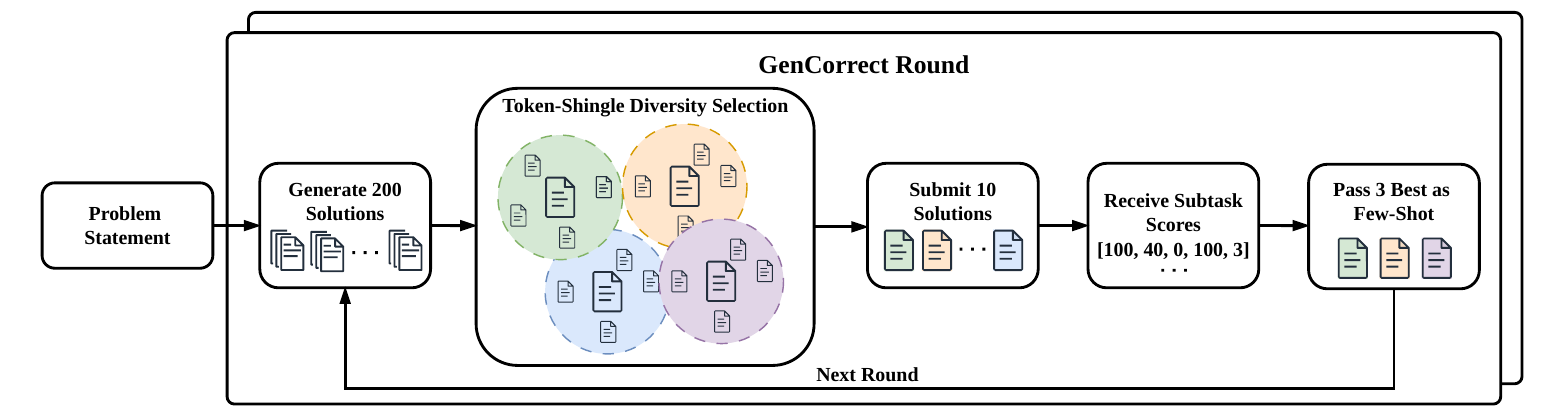}
    \caption{GenCorrect iteratively generates candidate solutions, selects a diverse subset for evaluation, and uses evaluator feedback to guide the next round.}
    \label{fig:gencorrect}
\end{figure*}

We introduce GenCorrect, an iterative test-time compute strategy applied for up to five rounds (Figure~\ref{fig:gencorrect}). GenCorrect combines large-scale sampling and behavior-based clustering for competitive programming~\citep{li2022alphacode,leblond2023alphacode2}, iterative self-critique and test-based refinement~\citep{ahmad2025opencodereasoningiisimpletesttime,ridnik2024codegenerationalphacodiumprompt}, and execution-grounded test-time selection~\citep{samadi2025gencluster,li2025stesttimescaling}. Each round consists of:

\begin{itemize}
    \item \textbf{Generation.}
    We generate up to 200 candidate solutions in parallel and compile them locally. The first round uses only the problem statement; subsequent rounds additionally use solutions and evaluator feedback from earlier rounds.

    \item \textbf{Diversity selection.}
    After filtering invalid outputs, we initialize a center set $C$ using a score-blind local heuristic $Q(c)$ and iteratively select the candidate farthest from the existing centers:
    \begin{equation}
    \label{eq:gencorrect_diversity}
        c_{\mathrm{next}}
        \in
        \arg\max_{c \notin C}
        \min_{z \in C}
        \left[1-\operatorname{sim}(c,z)\right].
    \end{equation}
     Here, $\operatorname{sim}(c,z)$ denotes the candidate similarity score.
  We select up to $K=10$ centers, assign every candidate to its most similar center, and choose the candidate with the
  highest $Q(c)$ as the representative of each cluster.

    \item \textbf{Execution.}
    We submit the 10 representatives to the evaluator. For IOI, feedback consists of subtask scores. All filtering and selection for the current round occur before these scores are observed.

    \item \textbf{Refinement.}
    We accumulate the best score observed for each subtask:
    \begin{equation}
    \label{eq:gencorrect_accumulation}
    \begin{aligned}
        A_r(t)
        &=
        \max\left(
            A_{r-1}(t),
            \max_{c\in S_r}s_t(c)
        \right),\\
        A_0(t)&=0,
    \end{aligned}
    \end{equation}
    where $S_r$ denotes the solutions submitted in round $r$. The next round is conditioned on the accumulated per-subtask score vector ($A_r$) and three complementary references selected to preserve solved subtasks, target remaining gaps, and maintain diversity.
\end{itemize}

For IOI, we perform five rounds of 10 submissions, matching the official limit of 50 submissions per problem. Previous works divide problems into subtasks while our approach provides all subtasks to the model and lets it decide which to work on, leading to substantially fewer necessary generations per problem \citep{openai2025competitive, samadi2025gencluster, yang2026nemotroncascade2}. For ICPC, where feedback is binary, we continue until the problem is solved or performance plateaus. Complete filtering, ranking, tie-breaking, and carry-forward rules along with the necessary prompts are provided in Appendix~\ref{app:gencorrect_details}.

\section{Experiments}
\subsection{Evaluation Setup}

We evaluate on IOI 2025, ICPC 2025, and LiveCodeBench Pro (LCB Pro)~\citep{ioi2025results,icpc2025results,zheng2025livecodebenchpro}, ensuring that all evaluation problems are excluded and deduplicated from our SFT and RL data. For IOI, Score@$k$ is computed by grouping generated solutions into independent runs. Within each run, we retain the highest score achieved on each subtask, sum across all subtasks and problems, and then average the resulting totals across runs. We report Score@1 for single-sample performance and Score@200 for parallel sampling. For ICPC and LCB Pro, Pass@1 is the fraction of problems solved by one sampled solution. IOI results are reported either as raw scores out of 600 or as normalized percentages, as indicated; ICPC Pass@1 is the percentage of the 12 problems solved. Final IOI and ICPC results are averaged over 1,000 runs, intermediate checkpoints over 50 runs, Score@200 over five runs, and LCB Pro over eight runs. Checkpoints are selected exclusively using the held-out validation set. We compare against gpt-oss-120b~\citep{openai2025gptoss}, Qwen3.6-35B-A3B~\citep{qwen2026qwen36}, Nemotron-Cascade 2~\citep{yang2026nemotroncascade2}, the base Nemotron-3 Nano and Ultra models~\citep{nvidia2025nemotron3nanoopen,nvidia2026nemotron3ultraopen}, DeepSeek-V4-Flash and DeepSeek-V4-Pro~\citep{deepseekai2026deepseekv4} (Max thinking), and GLM-5.2~\citep{glm5team2026glm5vibecodingagentic}. All reported competition results are obtained using our evaluation harness rather than copied from the corresponding model reports.

\subsection{Main Results}

\begin{figure*}[t]
    \centering
    \includegraphics[width=\linewidth]{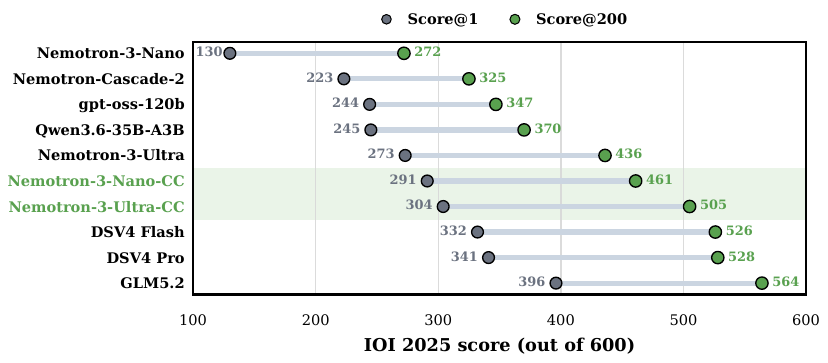}
    \caption{IOI 2025 Score@1 (grey) and Score@200 (green) across evaluated models with the grey line representing Score@k in between. Our submitted general Nemotron-3-Nano-CC and Nemotron-3-Ultra-CC is highlighted.}
    \label{fig:ioi_model_performance}
\end{figure*}

Figure~\ref{fig:ioi_model_performance} and Table~\ref{tab:main_results} summarize the main results while Figure~\ref{fig:combined_results} shows our Nano-CC progression with each pipeline stage. On IOI 2025, Nemotron-3-Nano-CC improves over its base model from 130 (21.7\%) to 291 (48.5\%) at Score@1 and from 272 to 461 at Score@200. Despite having only 3B active parameters, Nano-CC exceeds the base Nemotron-3-Ultra model at both sampling budgets. At Score@1, it outperforms all evaluated baselines except DeepSeek-V4-Flash, DeepSeek-V4-Pro, and GLM-5.2. The gains transfer beyond IOI: Nano-CC reaches 51.0\% Pass@1 on ICPC 2025 and 71.6\% on LCB Pro, compared with 16.9\% and 17.6\% for the base model. On LCB Pro, Nemotron-3-Nano-CC also exceeds gpt-oss-120b, Qwen3.6-35B-A3B, Nemotron-Cascade-2-30B-A3B, and DeepSeek-V4-Flash.

Nemotron-3-Ultra-CC, trained with SFT but without the CC RL stage, improves over the Ultra base model from 45.5\% to 50.7\% on IOI, 54.0\% to 57.4\% on ICPC, and 72.6\% to 74.5\% on LCB Pro. It exceeds Nano-CC by 2.2, 6.4, and 2.9 percentage points, respectively, and achieves the strongest results among our models on all three benchmarks. At its operating scale, Nano-CC is the strongest evaluated model with a comparable active parameter count, whereas Ultra-CC provides the highest absolute performance among our models even though it uses substantially less SFT training and does not feature RL.

\begin{table*}[t]
\centering
\small
\begin{tabularx}{\textwidth}{@{}>{\raggedright\arraybackslash}Xccc@{}}
\toprule
\textbf{Model} 
    & \shortstack{\textbf{IOI 2025}\\\textbf{Score@1}}
    & \shortstack{\textbf{ICPC 2025}\\\textbf{Pass@1}}
    & \shortstack{\textbf{LCB Pro}\\\textbf{Pass@1}} \\
\midrule

Nemotron-3-Nano-30B-A3B      & 21.7\% & 16.9\%     & 17.6\% \\
Nemotron-Cascade-2-30B-A3B          & 37.2\% & 42.0\%     & 45.6\% \\
gpt-oss-120b                         & 40.7\% & 45.8\%     & 66.4\% \\
Qwen3.6-35B-A3B                     & 40.8\% & 32.0\%     & 58.4\% \\
Nemotron-3-Ultra-550B-A55B           & 45.5\% & 54.0\%     & 72.6\% \\
DSV4 Flash                           & 55.3\% & 65.8\%     & 69.5\% \\
DSV4 Pro                             & 56.8\% & 69.6\%     & 78.2\% \\
GLM 5.2                             & 66.0\% & 65.7\%     & 83.8\% \\

\midrule

\textbf{Nemotron-3-Nano-CC}          & 48.5\% & 51.0\%     & 71.6\% \\
\textbf{Nemotron-3-Ultra-CC}         & 50.7\% & 57.4\%     & 74.5\%    \\
\bottomrule

\end{tabularx}
\caption{Main single-sample results on IOI 2025, ICPC 2025, and LiveCodeBench Pro. IOI Score@1 is reported as a normalized percentage, while ICPC and LiveCodeBench Pro report Pass@1.}
\label{tab:main_results}
\end{table*}

\newpage
\subsection{Effect of Supervised Fine-Tuning}

\begin{wrapfigure}{r}{0.50\textwidth}
\vspace{-4mm}
    \centering
    \includegraphics[width=\linewidth]
        {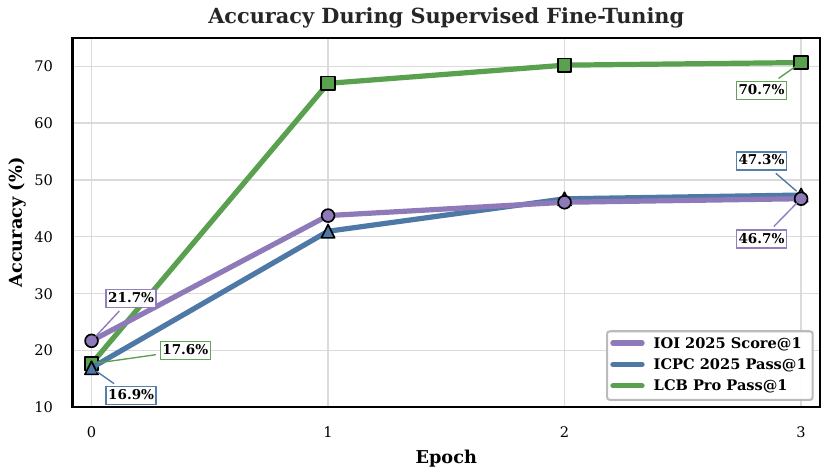}
    \caption{Normalized IOI 2025 Score@1, ICPC 2025 Pass@1, and LCB Pro Pass@1 before and during supervised fine-tuning of Nano-CC.}
    \label{fig:sft}
\end{wrapfigure}

Figure~\ref{fig:sft} shows that SFT produces most of Nano-CC's improvement. Over three epochs, IOI 2025 Score@1 increases from 21.7\% to 46.7\%, ICPC 2025 Pass@1 from 16.9\% to 47.3\%, and LCB Pro Pass@1 from 17.6\% to 70.7\%. Most gains occur during the first epoch, with performance beginning to saturate by the third. Ultra-CC receives a single SFT epoch over 477,642 examples, improving IOI from 45.5\% to 50.7\%, ICPC from 54.0\% to 57.4\%, and LCB Pro from 72.6\% to 74.5\%. These gains are smaller than Nano's SFT improvements of 24.8, 29.8, and 53.1 percentage points, respectively, reflecting the substantially stronger Ultra initialization. Nevertheless, the SFT-only Ultra-CC model outperforms the final Nano-CC model on all three benchmarks after roughly 478k SFT samples, whereas Nano-CC receives three SFT epochs 1.2M samples each, followed by RL. Thus, when model size and inference cost are not primary constraints, adapting a stronger base model with limited SFT can outperform extensive post-training of a smaller model.

\subsection{Effect of Reinforcement Learning}

\begin{figure*}[t]
    \centering
    \begin{subfigure}[t]{0.495\textwidth}
        \centering
        \includegraphics[width=\linewidth]
            {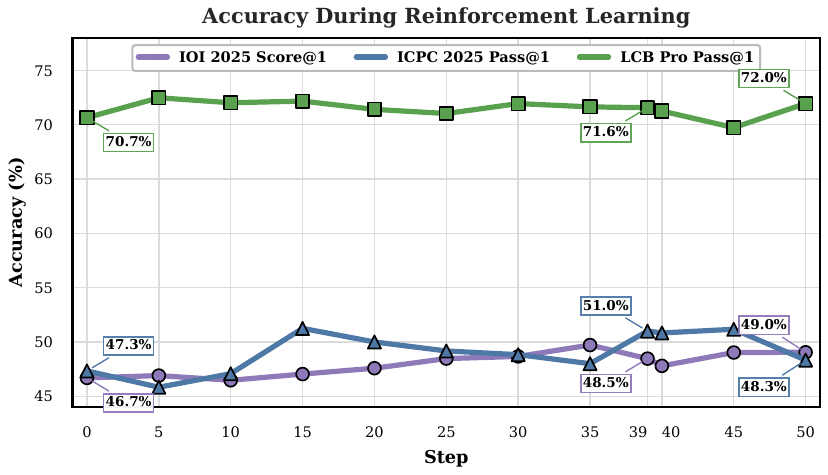}
        \caption{Performance during GRPO training. Step 39 is selected using held-out validation performance.}
        \label{fig:rl_performance}
    \end{subfigure}
    \hfill
    \begin{subfigure}[t]{0.495\textwidth}
        \centering
        \includegraphics[width=\linewidth]
            {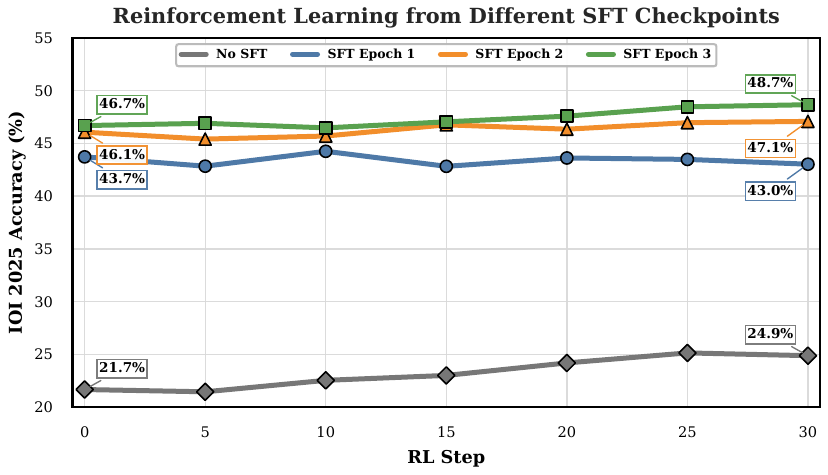}
        \caption{IOI 2025 Score@1 during RL when initialized from different SFT checkpoints.}
        \label{fig:rl_sft_checkpoint}
    \end{subfigure}
    \caption{Normalized performance during reinforcement learning of Nano-CC on multiple benchmarks and when initializing RL from different SFT checkpoints.}
    \label{fig:rl_results}
\end{figure*}

We apply RL only to Nano-CC, in part because of the substantial computational cost of running RL at Ultra's scale. As shown in Figure~\ref{fig:rl_performance}, starting from the third-epoch SFT checkpoint, RL improves IOI 2025 Score@1 from 46.7\% to 48.5\%, ICPC 2025 Pass@1 from 47.3\% to 51.0\%, and LCB Pro Pass@1 from 70.7\% to 71.6\%. Although smaller than the SFT gains, the improvements occur across all three benchmarks and we select step 39 exclusively using the held-out validation set.

We hypothesize that the modest gains reflect both the strong SFT initialization and the challenging optimization setting. With binary-reward GRPO, a problem provides a relative learning signal only when its rollout group contains both successful and unsuccessful solutions~\citep{shao2024deepseekmathpushinglimitsmathematical}, so RL primarily targets capabilities near the model's frontier. Furthermore, rollouts of up to 255K tokens receive only a terminal execution reward, creating a long-horizon credit-assignment problem. These long trajectories and sparse relative rewards may limit the magnitude and stability of the RL improvement.

Figure~\ref{fig:rl_sft_checkpoint} compares RL initialized from the base Nano checkpoint and from checkpoints after each SFT epoch. Without SFT, RL improves IOI 2025 Score@1 from 21.7\% to 24.9\% after 30 steps, demonstrating that executable-reward RL can produce measurable gains directly from the base model. However, RL does not recover the substantially larger gains provided by SFT due to the strength of the teacher models used: after 30 steps, models initialized from SFT epochs one, two, and three achieve 43.0\%, 47.1\%, and 48.7\%, respectively. Thus, under our training budget, RL can improve performance without prior SFT but does not substitute for the capabilities acquired through supervised fine-tuning, with the strongest final performance obtained from the third-epoch SFT checkpoint.

\subsection{Effect of Test-Time Compute}

\begin{figure*}[t]
    \centering

    \begin{subfigure}[t]{\textwidth}
        \centering
        \includegraphics[width=0.87\linewidth]
            {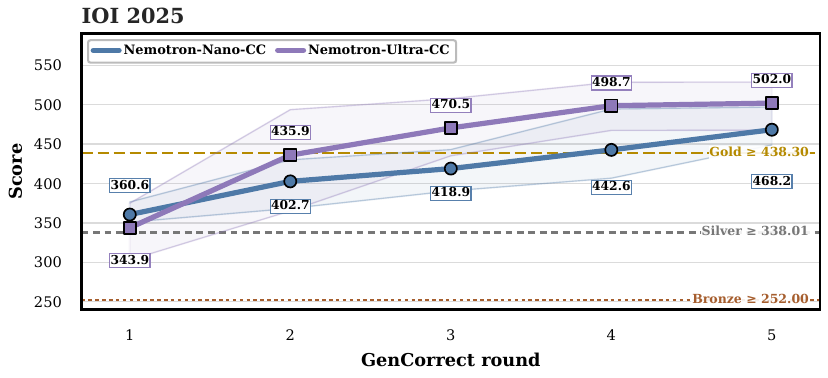}
        \caption{Mean IOI 2025 score across successive GenCorrect rounds for Nano-CC and Ultra-CC. Dashed lines indicate the medal thresholds, and shaded regions show the min--max range from 5 runs per round.}
        \label{fig:ioi_gencorrect_rounds}
    \end{subfigure}

    \vspace{0.3cm}

    \begin{subfigure}[t]{\textwidth}
        \centering
        \includegraphics[width=0.87\linewidth]
            {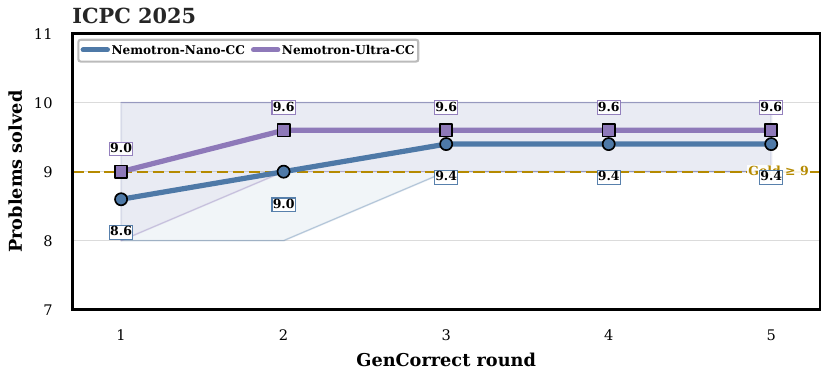}
        \caption{Mean ICPC 2025 problems solved across successive GenCorrect rounds for Nano-CC and Ultra-CC. Dashed line indicates the gold-medal threshold, and shaded regions show the min--max range from 5 runs per round.}
        \label{fig:icpc_gencorrect_rounds}
    \end{subfigure}

    \caption{Performance across successive GenCorrect rounds on (a) IOI 2025 and (b) ICPC 2025.}
    \label{fig:gencorrect_rounds}
\end{figure*}

Score@200 measures gains from parallel sampling, while GenCorrect additionally uses evaluator feedback to refine solutions across successive rounds and concentrate improvements from 200 generations into 10 submissions per round. As shown in Figure~\ref{fig:ioi_gencorrect_rounds}, Nano-CC's mean IOI 2025 score increases from 360.6 after the first round to 468.2 after five rounds, a gain of 107.6 points. Ultra-CC improves from 343.9 to 502.0, a substantially larger gain of 158.1 points. Although Ultra-CC begins below Nano-CC in the first round, it finishes 33.8 points ahead after five rounds. Relative to the official IOI 2025 thresholds, Ultra-CC exceeds the gold threshold after three rounds, while Nano-CC does so after four~\citep{ioi2025results}.

The larger GenCorrect improvement is consistent with a broader difference in how the two models benefit from test-time compute. At Score@1, Ultra-CC exceeds Nano-CC by only 2.2 percentage points, corresponding to approximately 13 raw IOI points. At Score@200, however, Ultra-CC reaches 505 compared with 461 for Nano-CC, widening the gap to 44 points. Thus, Ultra-CC's advantage grows substantially under parallel sampling and then is magnified by the iterative correction from GenCorrect. Together with its larger improvement across GenCorrect rounds, this suggests that Ultra-CC benefits both from stronger pool of candidate solutions from parallel sampling and from more effective use of evaluator feedback.

On ICPC 2025, Figure~\ref{fig:icpc_gencorrect_rounds} shows that the mean number of problems solved by Nano-CC increases from 8.6 to 9.4, reaching nine solved problems after two rounds and beginning to plateau after the third where nine solved problems match the result of the fourth-place gold-medal team~\citep{icpc2025results}. Ultra-CC starts at 9.0 problems solved and reaches 9.6 after two rounds, maintaining this performance through round five. It therefore reaches nine solved problems one round earlier than Nano-CC and remains ahead throughout the correction process. Both models plateau more quickly than on IOI, suggesting that ICPC's binary feedback provides less information for continued refinement than IOI's subtask-level scores.

\section{IOI 2026}
\label{sec_ioi_2026}
\subsection{Competition Setting}

We evaluate our system prospectively on the IOI 2026 problem set during the official competition and before the problems were publicly available. The competition comprised two five-hour sessions held over two days, with three problems released in each session. We operated under the same time, internet-access, and submission constraints as human contestants: internet access was prohibited, local code execution was permitted, and each problem allowed up to 50 submissions with one submission allowed per minute~\citep{ioi2026rules}. During the live inference deployment, we used a peak allocation of up to 760 NVIDIA GB300 GPUs.

\subsection{Competition-Specific Adaptations}

Having completed the general-pipeline experiments, we use IOI 2025 as a development benchmark and make several competition specific adaptations to maximize the score of a single live run within the official time and submission limits.

\paragraph{Ultra SFT with GLM-5.2 data.} Our previous results show that adapting the stronger Ultra model with limited SFT can outperform extensive post-training of the smaller Nano model, particularly in later GenCorrect rounds. We therefore focus on fine-tuning Ultra for the live IOI 2026 run. We use SFT rather than RL because SFT provides most of the post-training gains in our experiments, while RL at Ultra scale exceeds our available compute budget. We evaluate GLM-5.2 and DeepSeek-V4-Flash~\citep{glm5team2026glm5vibecodingagentic,deepseekai2026deepseekv4} on IOI 2025 as candidate SFT teachers. GLM-5.2 achieves a higher score with shorter outputs than DeepSeek-V4-Flash, as shown in Table~\ref{tab:ioi25_teacher_comparison}. This advantage transfers after SFT, with the GLM-5.2-trained Ultra-CC variant achieving both higher Score@1 and shorter average outputs than the DeepSeek-V4-Flash-trained variant. We therefore select GLM-5.2 training data for the live run.

\begin{table}[!htbp]
\centering
\begin{tabular}{lcc}
\toprule
\textbf{System}
& \textbf{IOI 2025 Score@1}
& \textbf{Mean Generation Length} \\
\midrule

GLM-5.2
& 66.0\% & 85,927 \\
DeepSeek-V4-Flash
& 55.3\% & 120,456 \\

\midrule

\textbf{Ultra-CC (GLM-5.2)}
& 59.4\% & 84,244 \\
\textbf{Ultra-CC (DeepSeek-V4-Flash)}
& 50.7\% & 89,626 \\

\bottomrule
\end{tabular}
\vspace{0.2cm}
\caption{IOI 2025 score and average number of generation tokens for the teacher models and corresponding Ultra-CC variants. Lower output length permits more candidates to be generated within a fixed inference window.}
\label{tab:ioi25_teacher_comparison}
\end{table}

\paragraph{Expanded final-round selection.}
For the first four GenCorrect rounds, we generate 200 solutions and submit 10 selected candidates per problem as described in Section \ref{subsection:ttc}. In the final round, we increase the generation budget to 1{,}000 solutions while retaining the final 10-submission limit. Inspired by GenCluster~\citep{samadi2025gencluster}, we use an execution-based selection procedure to rank this larger candidate pool:

\begin{enumerate}
\item We prompt the model to produce 50 problem-specific test-input generators and validators.
\item We execute and filter the generators using the validators until we obtain 100 valid test inputs.
\item We execute every compiled candidate solution on the generated inputs.
\item We prompt the model to produce a scoring script based on the problem's subtask criteria.
\item We use this script to rank the candidates and submit the 10 highest-ranked solutions.
\end{enumerate}

Because we have only a single live attempt, our objective is to maximize the score of that run using all available inference compute. Our IOI 2025 and ICPC 2025 results show that Ultra-CC begins to saturate over successive rounds of the general GenCorrect pipeline, suggesting limited benefit from using the standard 200-generation procedure again in the fifth round. We therefore use our available compute to expand the final-round candidate pool to 1{,}000 solutions. The execution-based selection procedure, adapted from GenCluster, enables us to select the most promising 10 submissions from this substantially larger pool and thereby make better use of the additional generation budget.

\paragraph{NVFP4 quantization.} To increase inference throughput, we apply post-training quantization to Nemotron-3-Ultra~\citep{nvidia2026nemotron3ultraopen} using the NVIDIA Model Optimizer NVFP4 recipe~\citep{nvidia-modelopt}. We calibrate the model on 1{,}000 sequences of 32{,}768 tokens sampled from our SFT mixture and use the resulting activation statistics to determine the quantization scales. The calibrated model is then exported to NVFP4 for live inference. Table~\ref{tab:nvfp4_quantization} reports the resulting trade-off between IOI 2025 performance and generation throughput. Across the evaluated NVFP4 configurations, IOI 2025 Score@1 remains within a narrow range of 52.7\%--53.5\%. In the matched BF16 KV-cache and prefix-enabled comparison, setting MTP to 5 nearly doubles throughput, from 345.9 to 698.5 tokens/s/GPU, while reducing Score@1 by only 0.6 percentage points, from 53.5\% to 52.9\%. For the live run, we select NVFP4 with an FP8 KV cache, prefix caching disabled, and MTP 5, which achieves 52.8\% Score@1 at 736.8 tokens/s/GPU. Relative to the BF16 baseline, this configuration sacrifices 6.6 percentage points of Score@1 for a 3.7$\times$ increase in throughput, enabling the large candidate batches required by GenCorrect within the competition window.

\begin{table}[!htbp]
\centering
\begin{tabular}{lcccccc}
\toprule
\textbf{Precision}
& \textbf{KV Cache}
& \textbf{Prefix}
& \textbf{MTP}
& \textbf{IOI 2025 Score@1}
& \shortstack{\textbf{Throughput}} \\
\midrule

BF16
& ---
& ---
& ---
& 59.4\%
& 199.1 \\

\midrule

NVFP4
& FP8
& Off
& 5
& 52.8\%
& 736.8 \\

NVFP4
& BF16
& Off
& 5
& 52.7\%
& 741.9 \\

NVFP4
& BF16
& On
& 5
& 52.9\%
& 698.5 \\

NVFP4
& BF16
& On
& Off
& 53.5\%
& 345.9 \\

\bottomrule
\end{tabular}
\vspace{0.2cm}
\caption{Effect of NVFP4 quantization and runtime configuration on IOI 2025 Score@1 performance and inference throughput. Throughput is normalized by the number of GPUs (tokens/s/GPU).}
\label{tab:nvfp4_quantization}
\end{table}

\subsection{Final Results}

Table~\ref{tab:ioi_2026} summarizes our IOI 2026 results, and Figure~\ref{fig:ioi_2026_result} visualizes the live result. Ultra-CC scores 535.4 during the competition window, exceeding the gold-medal threshold by 174.3 points and the top human contestant by 37.1 points~\citep{ioi2026results}. This result is obtained from a single prospective run using the competition-specific adaptations described above. To the best of our knowledge, this is the first time an AI system has outscored the highest-scoring human contestant on any IOI problem set. Moreover, our result was achieved live during IOI 2026, under the same time limits, submission platform, and internet restrictions as the human contestants.

To contextualize it against our general approach, we additionally conduct independent post-competition runs using the standard five-round GenCorrect pipeline. This pipeline achieves a mean score of 521.72, with an observed range of 495.0--545.8. The live result is 13.68 points above this mean, consistent with the intended benefit of the competition-specific adaptations, while remaining within the observed range. The general GenCorrect pipeline also achieves a mean score that exceeds both the gold-medal threshold and the highest human score.


\begin{table}[!htbp]
\centering
\begin{tabular}{lcc}
\toprule
\textbf{System} & \textbf{Score} & \textbf{Medal Level} \\
\midrule

Top Human Contestant
& 498.27 & Gold \\
Gold Medal Threshold
& 361.12 & Gold \\

\midrule

\textbf{Ultra-CC Live Competition Run}
& \textbf{535.40} & \textbf{Gold} \\


Ultra-CC General GenCorrect Pipeline
& 521.72 (495.0--545.8)
& Gold \\


\bottomrule
\end{tabular}
\vspace{0.2cm}
\caption{Performance on IOI 2026. The live competition result was
obtained under the same time and submission constraints as human
contestants. For the General GenCorrect Pipeline, we report the mean score followed by the observed minimum--maximum range across
5 independent runs.}
\label{tab:ioi_2026}
\end{table}

\section{Related Work}

Early work established competitive programming as a challenging evaluation setting for language models. AlphaCode combined domain-specific training with large-scale sampling, filtering, and behavioral clustering~\citep{li2022alphacode}, while AlphaCode 2 improved this approach using stronger models, additional fine-tuning, and learned reranking~\citep{leblond2023alphacode2}. LiveCodeBench and LiveCodeBench Pro subsequently introduced temporally separated and olympiad-level evaluations for measuring coding and reasoning performance~\citep{jain2025livecodebench,zheng2025livecodebenchpro}.

More recently, several systems have reported medal-level performance on international competitions. OpenAI's specialized o1-ioi system combined coding-focused reinforcement learning with a hand-engineered inference pipeline for IOI 2024, while o3 later exceeded the gold-medal threshold under the official submission limit~\citep{openai2025competitive}. OpenAI and Google DeepMind also reported gold-medal performance at ICPC 2025~\citep{icpc2025openai,lin2025geminiicpc}. Among open-weight models, GenCluster achieved IOI gold with an open-source model for the first time using large-scale generation and selection~\citep{samadi2025gencluster}, DeepSeek-V3.2-Speciale reported gold on IOI 2025 and ICPC 2025~\citep{deepseekai2025v32}, and Nemotron-Cascade 2 achieved gold-level performance with only 3B active parameters~\citep{yang2026nemotroncascade2}.

Our work also builds on research in synthetic reasoning data, execution-based reinforcement learning, and test-time scaling~\citep{snell2025scalingllmtesttimecompute}. OpenCodeReasoning demonstrated the effectiveness of synthetic reasoning traces for competitive programming~\citep{ahmad2025opencodereasoningadvancingdatadistillation}, while OpenCodeReasoning-II extended this direction with self-critique and iterative refinement~\citep{ahmad2025opencodereasoningiisimpletesttime}. Related approaches improve code generation at inference time through test-based iterative refinement~\citep{ridnik2024codegenerationalphacodiumprompt} and hybrid parallel and sequential scaling with execution-grounded selection~\citep{li2025stesttimescaling}. Prior work has also shown that self-correction without external feedback can be unreliable~\citep{huang2024largelanguagemodelsselfcorrect}, motivating the use of execution-grounded feedback during refinement. DeepSeek-R1 and DAPO established scalable reinforcement-learning recipes using verifiable rewards~\citep{guo2025deepseek,yu2025dapo}. Building on these directions, we study an end-to-end pipeline connecting synthetic data generation, supervised fine-tuning, reinforcement learning, and iterative test-time refinement.

\section{Conclusion}

We presented an end-to-end competitive-programming pipeline combining synthetic data, SFT, RL, and GenCorrect. Our experiments show that SFT provides the largest single-sample gains, RL provides smaller additional improvements, and GenCorrect substantially improves performance through feedback-driven refinement. On IOI 2025, Nano-CC improves from 130 to 468 points after post-training and GenCorrect, exceeding the gold threshold with 3B active parameters, while Ultra-CC reaches 502. Guided by these findings, we use IOI 2025 as a development benchmark to construct a competition-specific Ultra-CC system. The resulting system scored 535.4 out of 600 on the IOI 2026 problem set, exceeding both the gold threshold and the highest human score. To the best of our knowledge, this is the first time an AI system has outscored the highest-scoring human contestant on any IOI problem set. Moreover, our result was achieved live during IOI 2026, under the same time limits, submission platform, and internet restrictions as the human contestants.

\section*{Limitations}

Our approach requires substantial training and test-time compute. The live IOI result should therefore be interpreted as a system-level comparison under the same time and submission limits, rather than an equal-resource comparison with human contestants. Compute constraints also prevented RL training of Ultra-CC and exhaustive ablations across model scales and training stages. Our findings may not generalize beyond competitive programming. We plan to release our checkpoints and distributable inference and evaluation components through NeMo Skills. We cannot release the full training corpus because of third-party redistribution restrictions, but provide detailed data, training, and evaluation procedures.

\section*{Acknowledgments}

We are especially grateful to George Armstrong, Wei Du, and Igor Gitman for sharing their insights from IMO and contributing to our system, and to the IOI organization for their time and support.


{
\small
\bibliographystyle{references}
\bibliography{paper}
}

\appendix

\section{Data Curation Details}
\label{app:data_curation}

\paragraph{Problem collection.}
We collect 22,000 problems from 16 regional and international competition families spanning the last two decades, together with problems from online programming platforms. For each competition-year pair, an automated agentic pipeline retrieves the problem statement, time and memory limits, available official test cases, auxiliary files, and reference solutions. These components are packaged into a standardized executable evaluation environment.

\paragraph{Environment validation.}
We validate each environment by requiring known correct solutions to receive the expected score and known incorrect solutions to fail. As an additional validation step, we use gpt-oss-120b to generate candidate solutions, which are compiled and executed against the packaged test suite. This identifies malformed statements, missing auxiliary files, compiler incompatibilities, and inconsistent test harnesses that may not be exposed by the available reference solutions.

We remove problems whose environments produce inconsistent verdicts, whose reference solutions fail, or whose provided solutions do not compile. For the RL corpus, we additionally remove problems whose evaluation consistently requires more than 300 seconds when executed sequentially.

\paragraph{Evaluation contamination.}
We exclude all IOI 2025, ICPC 2025, and LiveCodeBench Pro problems from both the SFT and RL corpora and deduplicate the remaining training problems against these evaluations. IOI 2026 requires no retrospective contamination filtering because our system was executed during the live competition before the problems were publicly available.

\section{Training Details}
\label{app:training_details}

\subsection{Supervised Fine-Tuning}
\label{app:sft_details}

We use DeepSeek-V4-Flash to generate 1.2 million reasoning traces for Nemotron-3-Nano-30B-A3B and 477,642 traces for NVIDIA-Nemotron-3-Ultra-550B-A55B. We categorize problems as easy, medium, or hard and allocate additional generations to more difficult problems.

The SFT mixture also contains self-improvement traces in which the teacher receives a problem and a previously generated solution and is instructed to produce an improved solution. These examples expose the student models to the iterative refinement behavior used by GenCorrect at inference time.

Table~\ref{tab:sft_hyperparameters} reports the complete SFT configurations.

\begin{table*}[t]
    \centering
    \small
    \begin{tabular}{lcc}
        \toprule
        \textbf{Setting}
            & \textbf{Nano-CC}
            & \textbf{Ultra-CC} \\
        \midrule
        Initialization
            & Nemotron-3-Nano-30B-A3B
            & RLVR-teacher checkpoint \\
        Teacher model
            & DeepSeek-V4-Flash
            & DeepSeek-V4-Flash \\
        Training examples
            & 1,200,000
            & 477,642 \\
        Epochs
            & 3
            & 1 \\
        Optimizer
            & AdamW
            & AdamW \\
        Global batch size
            & 64
            & 64 \\
        Learning rate
            & $5 \times 10^{-5}$
            & $1.5 \times 10^{-5}$ peak \\
        Learning-rate schedule
            & Constant
            & Cosine \\
        Warmup ratio
            & 0
            & 0.1 \\
        Maximum packed sequence length
            & 262K
            & 262K \\
        Tensor parallelism
            & 4
            & 8 \\
        Context parallelism
            & 4
            & -- \\
        Hardware
            & 64 NVIDIA GB300-288GB GPUs
            & 128 NVIDIA GB300-288GB GPUs \\
        \bottomrule
    \end{tabular}
    \caption{Supervised fine-tuning configurations for Nano-CC and Ultra-CC.}
    \label{tab:sft_hyperparameters}
\end{table*}

Nano-CC achieves an average training throughput of 397.2 TFLOPs/s/GPU. Ultra-CC is initialized from its RLVR-teacher checkpoint, which was previously used to distill strong reasoning capabilities into the general-purpose model.

\subsection{Reinforcement Learning}
\label{app:rl_details}

\paragraph{Training and validation data.}
We apply RL only to Nemotron-3-Nano-30B-A3B. From the curated corpus, we initially identify 4,000 problems with reliable executable test cases. We remove problems whose evaluation consistently requires more than 300 seconds when executed sequentially, leaving 3,219 problems.

We divide these into 2,847 training and 372 validation problems. The split is performed at the parent-problem level so that subtasks derived from the same original problem cannot appear in both sets. To promote coverage across competitions, we target approximately one validation problem from each competition-year pair.

\paragraph{Rollout and reward configuration.}
We use Group Relative Policy Optimization (GRPO)~\citep{guo2025deepseek} with a maximum sequence length of 262,144 tokens, including up to 255,144 generated tokens. At each optimization step, we sample 16 rollouts at temperature 1.0 for each of 64 prompts, producing a global batch of 1,024 rollouts.

Each rollout generates a C++17 solution that is compiled and executed against the target problem or subtask. We assign a terminal reward of 1 when the solution receives full credit and 0 otherwise; partial scores do not produce intermediate rewards. Rewards are normalized within each rollout group, and relative advantages are computed using a leave-one-out baseline~\citep{ahmadian2024basicsrevisitingreinforcestyle}.

\paragraph{Optimization.}
We optimize a token-level clipped policy-gradient objective using Adam with a constant learning rate of $3 \times 10^{-6}$, zero weight decay, and no reference-policy KL penalty~\citep{yu2025dapo}. Generations truncated at the maximum context length are excluded from the policy loss.

The full RL configuration is summarized in Table~\ref{tab:rl_hyperparameters}.

\begin{table}[t]
    \centering
    \small
    \begin{tabular}{lc}
        \toprule
        \textbf{Setting} & \textbf{Value} \\
        \midrule
        Training problems & 2,847 \\
        Validation problems & 372 \\
        Algorithm & GRPO \\
        Prompts per step & 64 \\
        Rollouts per prompt & 16 \\
        Rollouts per step & 1,024 \\
        Sampling temperature & 1.0 \\
        Maximum sequence length & 262,144 \\
        Maximum generation length & 255,144 \\
        Reward & Binary full credit \\
        Optimizer & Adam \\
        Learning rate & $3 \times 10^{-6}$ \\
        Learning-rate schedule & Constant \\
        Weight decay & 0 \\
        Reference-policy KL & None \\
        \bottomrule
    \end{tabular}
    \caption{Reinforcement-learning configuration for Nano-CC.}
    \label{tab:rl_hyperparameters}
\end{table}

\paragraph{Checkpoint selection.}
We evaluate intermediate checkpoints on the held-out validation set. The checkpoint at step 39 achieves the highest validation performance and is used for all subsequent evaluations.

\section{Additional GenCorrect Details}
\label{app:gencorrect_details}

\paragraph{Candidate filtering.}
We extract and normalize the C++ program from each model response before computing program features or similarities. Before clustering, we retain candidates that compile, contain nonempty normalized code with at least 40 tokens, and have a nonnegative structural-feature score. If no candidates satisfy all requirements, we progressively relax them until the candidate pool is nonempty.

\paragraph{Similarity and clustering.}
The similarity function $\operatorname{sim}(c,z)$ in Equation~\ref{eq:gencorrect_diversity} compares token shingles extracted from the normalized programs. We initialize the center set with the highest-ranked candidate under the score-blind heuristic $Q(c)$. Subsequent centers are selected using Equation~\ref{eq:gencorrect_diversity}, with ties broken by $Q(c)$. After selecting $K=10$ centers, each remaining candidate is assigned to its most similar center. Equal-similarity ties are assigned to the smaller cluster, with any remaining ties resolved by center-selection order.

\paragraph{Score-blind ranking.}
The heuristic used for center initialization, tie-breaking, and representative selection is

\begin{equation}
\resizebox{\linewidth}{!}{$
\begin{aligned}
Q(c)=\big(
    &\operatorname{Compiles}(c),
    \operatorname{FeatureScore}(c),
    \lfloor\operatorname{CodeLength}(c)/800\rfloor,\\
    &\operatorname{ExactFrequency}(c),
    \operatorname{CodeLength}(c),
    -\operatorname{GeneratedTokens}(c),
    -\operatorname{RunId}(c)
\big),
\end{aligned}
$}
\label{eq:gencorrect_heuristic}
\end{equation}

with entries compared lexicographically and larger values preferred. Here, $\operatorname{ExactFrequency}(c)$ is the number of candidates with the same normalized code, $\operatorname{GeneratedTokens}(c)$ is the length of the original model output, and $\operatorname{RunId}(c)$ is its deterministic generation index.

The feature score rewards indicators of complete C++ programs, including headers, a main function, input/output operations, and a return statement, while penalizing unfinished placeholders. Neither $Q(c)$ nor the clustering procedure uses competition scores.

\paragraph{Representative selection.}
We select the highest-ranked candidate under $Q(c)$ from each cluster. If this produces fewer than 10 representatives, the remaining submission slots are filled with the highest-ranked unselected candidates. Thus, cluster membership promotes diversity, while $Q(c)$ selects a locally well-formed representative from each cluster. All filtering, clustering, and representative selection for a round are completed before its evaluator feedback is observed.

\paragraph{Feedback and carry-forward.}
The accumulated subtask scores $A_r(t)$ in Equation~\ref{eq:gencorrect_accumulation} form an internal feedback state used to guide subsequent generations; they do not alter the evaluator's official scoring procedure.

After each round, the highest-total-scoring submission from that round becomes the primary solution. Ties are resolved using the score-blind heuristic. We then select three complementary references from the current submissions and solutions retained from previous rounds:

\begin{enumerate}
    \item the strongest overall candidate across the available solutions;
    \item a candidate addressing the largest remaining subtask gap; and
    \item a candidate with strong aggregate coverage of the remaining gaps.
\end{enumerate}

Subtask gaps are determined from the accumulated feedback state $A_r$. Similarity and $Q(c)$ are used as tie-breakers to avoid redundant references and prefer locally well-formed programs. The next round is conditioned on the problem statement, the accumulated subtask feedback, the primary solution, and these three references.

\newpage
\subsection{Test-Time Compute Prompts}

\begin{tcolorbox}[enhanced, breakable, title={First Gencorrect Round}, colback=red!0, left=2pt, right=2pt, top=2pt, bottom=2pt]
{ 
\begin{verbatim}
system: |-
  Your response must use the following format:

  Explanation: <your explanation for the final answer>
  Confidence: <your confidence score between 0% and 100%>
  Answer:
  ```cpp
  <your complete C++20 implementation>
  ```

  The Answer must contain exactly one complete C++20 code block, and that
  code block must be the final content in your response. Do not write
  anything after its closing fence.

user: |-
   You are an expert competitive programmer. You will be given a problem
   statement, test case constraints, and example test inputs and outputs.
   Please reason step by step about the solution, then provide a complete
   implementation in C++20.

   You should correctly implement the routine(s) described in the
   Implementation Details, without reading or writing anything directly
   from stdin or to stdout, as input and output are passed through the
   implemented routines.

   Assume your code will be run on the OFFICIAL grader, and do not add a
   main function, a sample grader, or any other functionality unless it
   has been explicitly requested.

   If multiple subtasks are listed, first choose one subtask to target,
   state that choice briefly in your reasoning, and focus on an approach
   that is sufficient for that chosen subtask rather than trying to solve
   every listed subtask.

   Put your final solution within a single code block:

   ```cpp
   <your code here>
   ```

   {problem}
\end{verbatim}

}
\end{tcolorbox}

\newpage

\begin{tcolorbox}[enhanced, breakable, title={Next Gencorrect Round}, colback=red!0, left=2pt, right=2pt, top=2pt, bottom=2pt]
{\begin{verbatim}
system: |-
  Your response must use the following format:

  Explanation: <your explanation for the final answer>
  Confidence: <your confidence score between 0% and 100%>
  Answer:
  ```cpp
  <your complete C++20 implementation>
  ````

  The Answer must contain exactly one complete C++20 code block, and that
  code block must be the final content in your response. Do not write
  anything after its closing fence.

user: |-
  You are an expert competitive programmer.

  You will be given:

  * the problem statement
  * multiple candidate solutions from previous evaluated generations
  * each candidate solution's subtask scores
  * the best achieved subtask scores so far across rounds
  * the maximum subtask scores

  The candidate solutions are provided only as inspiration.
  They may solve different subtasks, may contain bugs, and may be
  incomplete. There is no main candidate solution; treat all candidate
  solutions as peer references. You may discard them completely and
  write a new solution from scratch.

  Choose one target subtask using the best achieved scores so far:

  * a subtask is eligible only if 
    achieved_subtask_scores[subtask] < max_subtask_scores[subtask]
  * choose exactly one eligible target subtask and focus on solving 
    that target better
  * prefer the eligible target with the largest remaining score gap, 
    unless another eligible target is clearly easier to improve
  * do not spend effort on subtasks that are already fully achieved

  Use the candidate solutions to identify useful ideas, edge cases, or
  implementation patterns. Do not simply concatenate candidate solutions.
  Do not spend time reconciling differences between candidate solution
  scores and best achieved scores so far. Do not choose a target based on
  any single candidate's scores alone.

  If all candidate solutions mainly solve already-achieved subtasks,
  ignore them and design a fresh approach for the chosen unsolved target.

  In your reasoning:

  1. identify which subtasks are still globally unsolved or partially 
     solved using achieved_subtask_scores and max_subtask_scores
  2. choose exactly one target subtask
  3. briefly identify any useful ideas from the candidate solutions, if any
  4. explain an approach sufficient for the chosen target
  5. then provide a complete C++20 implementation

  You should correctly implement the routine(s) described in
  Implementation Details, without reading or writing anything directly
  from stdin or to stdout, as input and output are passed through the
  implemented routines.

  Assume your code will be run on the OFFICIAL grader, and do not add a
  main function, a sample grader, or any other functionality unless it
  has been explicitly requested.

  Put your final solution within a single code block:

  ```cpp
  <your code here>
  ```

  ## Problem
  {problem}
 
  ## Candidate Solutions From Previous Evaluated Generations
  {candidate_solutions}

  ## Best Achieved Subtask Scores So Far
  {achieved_subtask_scores}

  ## Maximum Subtask Scores
  {max_subtask_scores}
\end{verbatim}

}
\end{tcolorbox}

\end{document}